\documentclass{article} 
\usepackage{iclr2027_conference,times}

\usepackage{amsmath,amsfonts,bm}

\def\eqref#1{equation~\ref{#1}}

\def\1{\bm{1}}

\DeclareMathAlphabet{\mathsfit}{\encodingdefault}{\sfdefault}{m}{sl}
\SetMathAlphabet{\mathsfit}{bold}{\encodingdefault}{\sfdefault}{bx}{n}

\usepackage{hyperref}
\hypersetup{hidelinks}
\usepackage{url}\usepackage{graphicx}
\usepackage{booktabs}
\usepackage{amssymb}
\usepackage{colortbl}
\definecolor{gpubrow}{RGB}{232,241,248}
\definecolor{refblue}{rgb}{0.88,0.94,1.00}
\definecolor{promptborder}{rgb}{0.68,0.74,0.82}
\definecolor{promptbg}{rgb}{0.965,0.975,0.988}
\usepackage{multirow}
\usepackage{float}
\usepackage{placeins}

\title{\centering Grounded Product Understanding\\in Livestream Videos}

\author{%
\makebox[\dimexpr\textwidth-2\tabcolsep\relax][c]{%
\begin{tabular}{c}
\textbf{Xinyu Zhang}\textsuperscript{1},
\textbf{Junjie Chen}\textsuperscript{1},
\textbf{Jiawei Ge}\textsuperscript{2}\thanks{Corresponding author. E-mail: \href{mailto:0jiawei.ge0@gmail.com}{\texttt{0jiawei.ge0@gmail.com}}.},
\textbf{Qianlong Li}\textsuperscript{1}\\
\textbf{Libin Ma}\textsuperscript{1},
\textbf{Baokun Pan}\textsuperscript{1}
\& \textbf{Yahui Luo}\textsuperscript{1}\\[2pt]
{\normalfont\textsuperscript{1}Kuaishou Technology
\qquad
\textsuperscript{2}Institute of AI for Industries, Chinese Academy of Sciences}
\end{tabular}}%
}

\iclrfinalcopy
\begin{document}
\maketitle
\pagestyle{plain}
\thispagestyle{plain}
\begin{abstract}
E-commerce livestreams have emerged as an important channel for presenting products to online consumers, often featuring multiple products with relevant information distributed across different moments. This poses significant challenges for downstream product understanding applications, such as product-centric livestream clipping, where models need to identify the product and its relevant segments for information gathering. However, existing benchmarks for general product understanding typically evaluate product retrieval and temporal localization in isolation, leaving the critical correspondence between product identity and temporal evidence largely unassessed. To address this limitation, we introduce \textbf{GPUB}, a large-scale benchmark comprising 3,000 real-world e-commerce livestream instances with quality-controlled multi-moment temporal annotations and a catalog of over 31K fashion products. GPUB supports three evaluation tasks: given a livestream video and a candidate product set, the main task \textbf{Grounded Product Understanding (GPrU)} requires jointly identifying the product being presented and localizing its supporting moments; Product Retrieval and Product Moment Localization serve as two complementary subtasks. Evaluation of existing multimodal models shows that GPrU remains highly challenging, with the best-performing off-the-shelf baseline achieving only 10.13\% Pair mAP@.3. To narrow the performance gap, we further develop UniPro,  a unified product understanding model that derives product-aligned and temporally structured representations from shared multimodal encoding, improving Pair mAP@.3 to 24.58\% while achieving 38.81\% Joint R@1@.3 on GPrU.
\end{abstract}
\section{Introduction}
\label{sec:introduction}

\begin{figure}[t!]
\centering
\includegraphics[width=\linewidth]{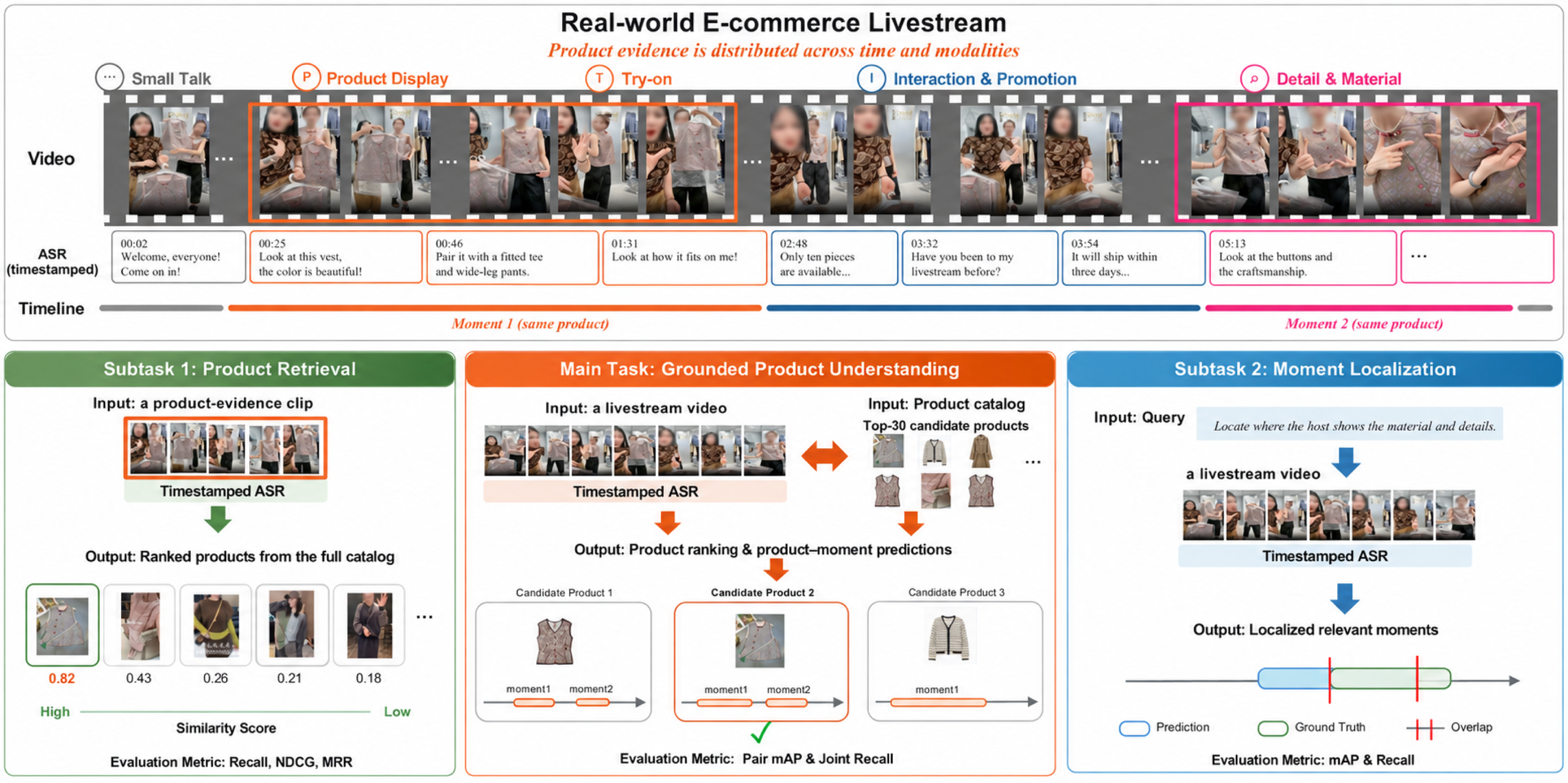}
\caption{Grounded Product Understanding in e-commerce livestreams. Top: An example e-commerce livestream with product evidence distributed across disjoint moments, conveyed through both visual content and timestamped ASR. Bottom: The three evaluation tasks in GPUB.}
\label{fig:gpub_task_definition}
\end{figure}

Livestream e-commerce enables hosts to showcase and sell products through continuous video streams with real-time viewer interaction. Unlike information-dense product pages or edited short videos, product evidence is distributed across multiple non-contiguous moments, with visual and spoken cues providing complementary information (Fig.~\ref{fig:gpub_task_definition}). This fragmented and multimodal evidence makes establishing reliable correspondence between catalog products and their supporting livestream moments inherently challenging, creating a key bottleneck for downstream applications such as product-centric clipping and summarization.

Existing tasks address product identification or temporal localization from different perspectives, but largely treat them as separate problems. E-commerce video benchmarks evaluate general product understanding through question answering~\citep{liu2026evads}, while product representation and retrieval methods focus on matching livestreams or short videos to specific products~\citep{bai2023crossdomain,yang2023crossview,hu2024sgmn}. Product retrieval focuses on resolving which catalog product is presented; however, it operates on temporally trimmed or product-centric video inputs, where the relevant content has already been narrowed to some extent. Temporal grounding, in contrast, explicitly evaluates when relevant evidence occurs~\citep{hendricks2017localizing,lei2021qvhighlights,liu2025timescope}, yet typically assumes that the target semantics are already specified by a text query, question, or task description. In general, neither setting requires the target product and its supporting moments to be inferred jointly from the same livestream video. As a result, a model may perform well on both aspects independently without establishing the correct correspondence between the identified product and its supporting moments.

To address this gap, we formulate a joint problem as \textbf{Grounded Product Understanding (GPrU)}, requiring jointly identifying the target catalog product and localizing its supporting moments from the same livestream video. We further introduce \textbf{Grounded Product Understanding Benchmark (GPUB)} for this task, containing 3,000 livestream evaluation windows, 8,975 quality-controlled product evidence moments, and a catalog of 31,831 fine-grained fashion products. GPUB defines Grounded Product Understanding as the main task, alongside Product Retrieval and Product Moment Localization as two complementary subtasks (Fig.~\ref{fig:gpub_task_definition}). These evaluations assess product identification and temporal localization separately, as well as the correspondence between identified products and their supporting moments in GPrU.

GPUB combines three key design features to support these evaluations: (1) a fine-grained fashion product catalog that consists entirely of apparel categories and captures the challenge of distinguishing highly similar fashion products; (2) temporally distributed multimodal evidence annotations that require models to integrate visual and spoken clues across multiple moments; and (3) an identity-preserving joint annotation and evaluation protocol that ensures product identification and temporal localization refer to the same product.

To provide a domain-trained baseline for GPUB, we develop UniPro, a unified model designed to capture both product-level semantics and temporal evidence. Fine-grained product identification emphasizes semantic discrimination between similar items, while evidence localization additionally requires temporal structure. UniPro addresses this through shared multimodal encoding, then derives a product-aligned representation for product identification and temporally structured representations for evidence localization. Our main contributions are as follows:
\begin{itemize}
    \item We introduce Grounded Product Understanding (GPrU), a new task that requires jointly identifying the target catalog product and localizing its supporting semantic evidence.
    \item We construct GPUB, a large-scale benchmark to support product understanding in e-commerce livestreams, with multi-moment evidence annotations and a fine-grained fashion product catalog. GPUB evaluates GPrU as its main task, with Product Retrieval and Product Moment Localization serving as two complementary subtasks.
    \item Extensive evaluation of existing off-the-shelf MLLMs reveals substantial limitations in modeling product--moment correspondence. We further develop UniPro, a unified product understanding model, as a task-trained solid baseline to facilitate future research on GPrU.
\end{itemize}

\section{Related Work}
\label{sec:related}

\paragraph{E-commerce and Livestream Product Retrieval.} Product retrieval aims to identify corresponding items from a product catalog, often requiring fine-grained discrimination among similar products. Early work primarily studies image--text matching over large product collections~\citep{zhan2021product1m}. With the growing prevalence of video-based shopping, subsequent studies extend product retrieval to short videos and livestreams, addressing challenges such as temporal variation, cross-domain appearance shifts, and background distractors~\citep{zhang2021fashionfocus,yang2023crossview,hu2024sgmn}. However, existing work largely focuses on product matching and exhibition intervals, rather than associating products with moments that convey related product information.

\paragraph{Video Temporal Grounding and Moment Retrieval.}
Video temporal grounding aims to localize video segments corresponding to a given language query. Existing methods span proposal-based matching, direct boundary prediction, and unified temporal modeling~\citep{hendricks2017localizing,zhang2020vslnet,lei2021qvhighlights,moon2023qddetr,lin2023univtg}. Recent work has further broadened temporal grounding to diverse settings, including long-form videos~\citep{soldan2022mad,tan2024synopground}, grounded question answering~\citep{xiao2024nextgqa}, task-oriented evidence localization~\citep{liu2025timescope}, and MLLM-based temporal grounding~\citep{zhang2026timelens,zheng2026taro}. Despite these extensions, existing temporal grounding methods typically assume that the target is directly specified by a query, whereas GPrU requires the target product to be inferred from a candidate set.

\paragraph{Long-form Multimodal Video Modeling.} Long-form video understanding is constrained by the amount of visual and linguistic context that can be processed effectively. Existing approaches address the visual side through adaptive frame selection, token compression, and memory-based summarization, enabling models to cover longer temporal contexts~\citep{shen2024longvu,buch2025flexible,hu2025frameselection,
song2024moviechat,he2024malmm}. Complementary work incorporates subtitles or ASR to provide semantic information beyond visual frames, ranging from video--language pretraining~\citep{li2020hero,xu2021videoclip} to streaming multimodal modeling~\citep{chen2025livecc}. These approaches enable more effective long-form video understanding, while product understanding in e-commerce livestreams requires identifying information relevant to specific products from complex visual scenes and semantically noisy spoken content.

\section{The GPUB Benchmark}
\label{sec:benchmark}

 This section presents the task formulations, data construction pipeline, dataset characteristics, and evaluation protocols of GPUB.

\subsection{Task Definition}
\label{sec:task_definition}

Let $\mathcal{G}=\{p_1,\ldots,p_{N_g}\}$ denote the shared product catalog.
A benchmark instance consists of a livestream video $V$ centered on a single target product $p^*\in\mathcal{G}$,
its timestamped ASR $A$,
and a set of ground-truth product evidence moments
$\mathcal{E}^*$. Each $e=[t_s,t_e]\in\mathcal{E}^*$ is a temporal interval
that provides evidence for $p^*$. GPUB focuses on three tasks:
\textbf{(1) Product Retrieval (subtask).} Given a product evidence clip
$(V_m,A_m)$ cropped from $(V,A)$ and the full catalog $\mathcal{G}$, the model
ranks catalog products by their correspondence to the clip.
\textbf{(2) Product Moment Localization (subtask).} Given the full video
$(V,A)$ and a product-related text query $q$, the model returns confidence-ranked temporal intervals that support the semantics of $q$. \textbf{(3) Grounded Product Understanding (main task).} Given the full instance video $(V,A)$ and a fixed candidate set $\mathcal{C}\subset\mathcal{G}$ of 30 products containing $p^*$, with no text query provided, the model ranks the candidate products and produces product--moment predictions $\{(p_i,\hat m_{ij})\}$, where $p_i\in\mathcal{C}$ and $\hat m_{ij}=[\hat t_s^{ij},\hat t_e^{ij}]$ denotes the predicted moment corresponding to $p_i$.

\subsection{Dataset Construction}
\label{sec:dataset_construction}

\begin{figure}[t!]
    \centering
    \includegraphics[width=0.92\linewidth]{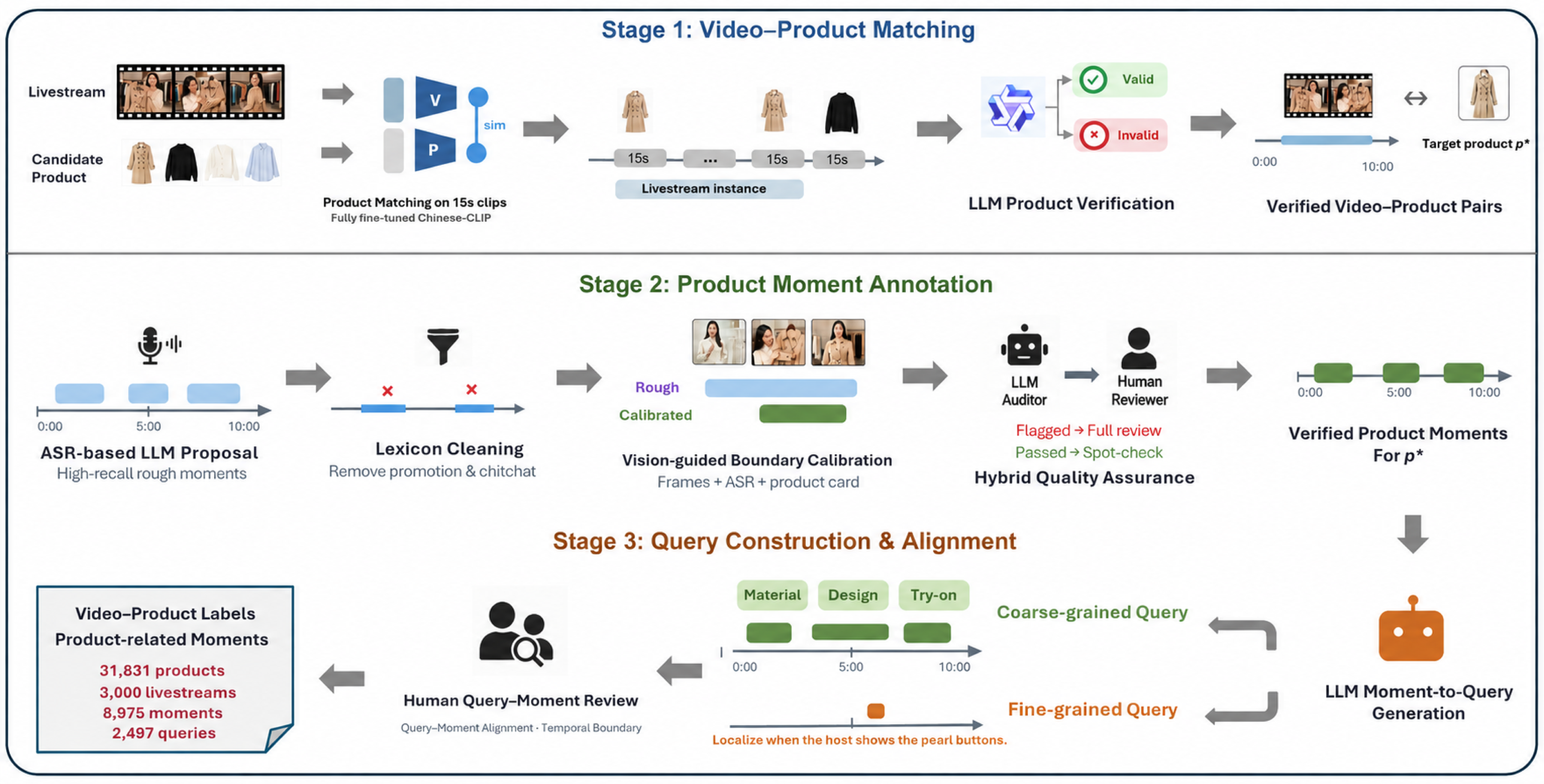}
    \caption{Video-side construction pipeline. \textbf{Stage 1} verifies
    video--product pairs. \textbf{Stage 2} annotates the complete set of
    evidence corresponding to each target product. \textbf{Stage 3} derives
    coarse-grained and fine-grained queries and aligns them with their
    supporting moments.}
    \label{fig:dataset_construction}
\end{figure}

We construct GPUB by progressively establishing the target product and its
supporting temporal evidence for each livestream. Specifically, we first curate the product
catalog by normalizing product identifiers, merging duplicate SKUs,
standardizing product metadata, and retaining one quality-controlled main image
for each product, resulting in a catalog of 31,831 products. Building on this catalog, the video-side pipeline comprises three stages (Figure~\ref{fig:dataset_construction}): video–product matching, product evidence annotation, and query construction and evidence alignment. The first two stages provide the joint annotations for Grounded Product Understanding, while the third derives textual queries and their query-specific temporal annotations for Product Moment Localization.

\paragraph{Stage 1: Video--product matching.}
We first determine the specific catalog product being discussed in the
livestream instance. A fully SFT-trained Chinese-CLIP model retrieves candidate products from each 15-second clip, after which Qwen3-VL-32B-Instruct verifies the top-ranked candidate using the entire instance video, timestamped ASR, and product information. We retain only
instances for which the product match can be confidently verified and discard
uncertain cases. The verified product is taken as the target product $p^*$,
which provides the basis for subsequent temporal evidence annotation.

\paragraph{Stage 2: Product evidence annotation.}
We annotate $\mathcal{E}$, the set of \emph{product evidence
moments}: temporally coherent intervals that refer to $p$
and convey visual or spoken information about its attributes
or characteristics. Non-contiguous evidence is annotated separately, while intervals without substantive product information, such as those involving promotion, pricing, inventory updates, or casual conversation, are excluded.
Qwen3-VL-235B-A22B-Instruct~\citep{bai2025qwen3vl}, with
thinking enabled, first proposes high-recall candidate intervals
from timestamped ASR, followed by lexicon-based screening
of promotional and conversational noise. The model then
validates the remaining candidates using video frames, ASR,
and target product information, and calibrates their temporal
boundaries against the visual content.
GPT-4o~\citep{openai2024gpt4o} flags uncertain annotations
for full human review, while a random subset of passed instances is fully audited to verify annotation correctness and identify missing evidence.

\paragraph{Stage 3: Query construction and evidence alignment.}
Given $\mathcal{E}^*$, GPT-4o~\citep{openai2024gpt4o} constructs
queries and their temporal targets for Product Moment
Localization at two semantic granularities.
A \emph{coarse-grained query} covers a broad aspect, such as
material, design, or try-on experience, and is aligned with
the verified moments conveying that information.
A \emph{fine-grained query} targets a specific attribute or
demonstration, such as fabric texture or lace-paneling details,
with boundaries further refined to precisely cover the
supporting evidence. Human reviewers then validate each query–evidence pair and refine temporal boundaries when necessary, yielding the final correspondences $(q,\mathcal{E}_q)$.

\paragraph{30-candidate construction.}
To focus GPrU on joint product identification and temporal grounding rather than large-scale retrieval, we construct a challenging 30-product
candidate set for each instance. We use Qwen3-VL-Embedding-2B~\citep{li2026qwen3vlembedding}
to retrieve the 29 highest-ranked non-target products from the full catalog
as hard negatives. Together with the target product \(p^*\), they form the
candidate set, with \(p^*\) randomly positioned
to avoid positional bias. GPrU
is evaluated over these candidate sets, whereas Product Retrieval is evaluated over the full catalog.

\begin{table*}[t]
    \centering
    \caption{Comparison of representative benchmark test splits. Counts refer to the evaluation split where reported; -- indicates data not reported or not released for that split. \(\dagger\) denotes clothing-trajectory queries (Video2Shop), cropped clips (LPR4M), and query--video pairs (QVHighlights), rather than distinct source videos. ASR indicates whether speech transcripts are provided with the evaluation data; QVHighlights releases ASR captions for training videos only.}
    \label{tab:benchmark_comparison}
    \scriptsize
    \setlength{\tabcolsep}{2.6pt}
    \renewcommand{\arraystretch}{0.94}
    \begin{tabular*}{\textwidth}{@{\extracolsep{\fill}}llrrrccccc@{}}
        \toprule
        \multirow{2}{*}{\textbf{Benchmark}}
        & \multirow{2}{*}{\textbf{Domain}}
        & \multicolumn{4}{c}{\textbf{Dataset properties}}
        & \multicolumn{4}{c}{\textbf{Evaluation coverage}} \\
        \cmidrule(lr){3-6}\cmidrule(lr){7-10}
        & & \shortstack{\textbf{Catalog}\\\textbf{products}}
        & \textbf{Videos}
        & \shortstack{\textbf{Avg.}\\\textbf{duration}}
        & \textbf{ASR}
        & \shortstack{\textbf{Product}\\\textbf{ID}}
        & \shortstack{\textbf{Temporal}\\\textbf{loc.}}
        & \shortstack{\textbf{Multi-}\\\textbf{moment}}
        & \shortstack{\textbf{Joint}\\\textbf{P--M}} \\
        \midrule
        Video2Shop~\citep{cheng2017video2shop} & Fashion & -- & 5,266\textsuperscript{$\dagger$} & -- & --
        & \(\checkmark\) & -- & -- & -- \\
        LPR4M~\citep{yang2023crossview} & \shortstack[l]{Livestream\\commerce} & 66,358 & 20,079\textsuperscript{$\dagger$} & -- & \(\checkmark\)
        & \(\checkmark\) & -- & -- & -- \\
        Charades-STA~\citep{gao2017tall} & \shortstack[l]{Indoor\\activities} & -- & 1,334 & 30.6\,s & --
        & -- & \(\checkmark\) & -- & -- \\
        QVHighlights~\citep{lei2021qvhighlights} & Vlogs/news & -- & 1,542\textsuperscript{$\dagger$} & $\approx$150\,s & --
        & -- & \(\checkmark\) & \(\checkmark\) & -- \\
        MomentSeeker~\citep{yuan2025momentseeker} & Open-domain & -- & 268 & 1,201.9\,s & --
        & -- & \(\checkmark\) & \(\checkmark\) & -- \\
        \midrule
        \rowcolor{gpubrow}
        \textbf{GPUB (ours)} & \shortstack[l]{\textbf{Fashion}\\\textbf{livestream}} & \textbf{31,831} & \textbf{3,000} & \textbf{265.1\,s} & \(\checkmark\)
        & \(\checkmark\) & \(\checkmark\) & \(\checkmark\) & \(\checkmark\) \\
        \bottomrule
    \end{tabular*}
\end{table*}
\subsection{Benchmark Characteristics}
\label{sec:benchmark_statistics}

\begin{figure*}[t]
    \centering
    \includegraphics[width=\textwidth]{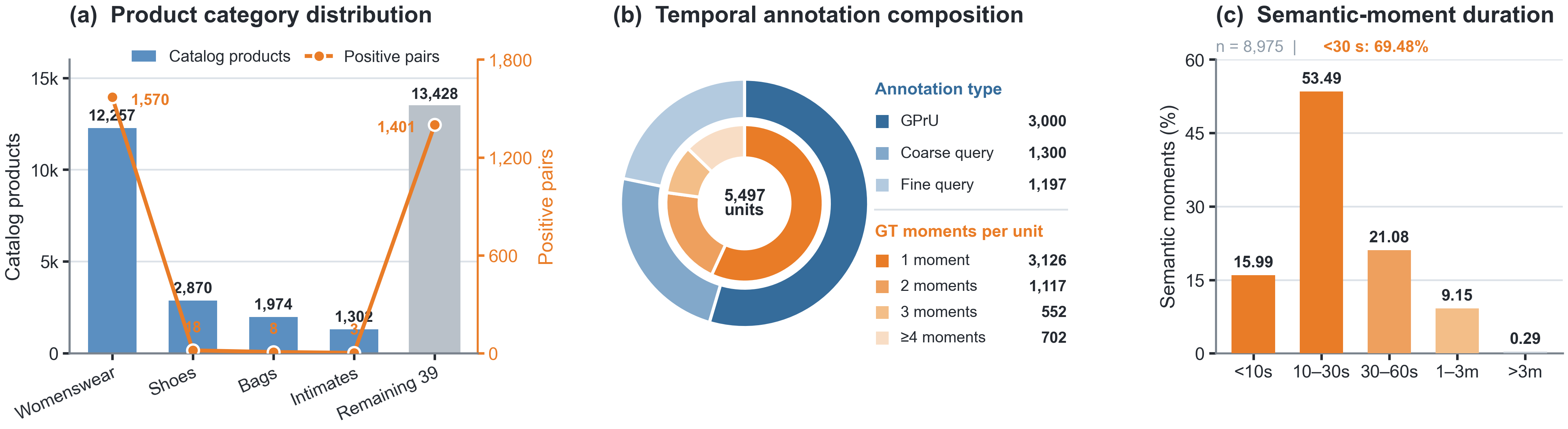}
    \caption{GPUB characteristics. \textbf{(a)} Distribution of catalog products and positive pairs across major product categories, with the remaining 39 categories aggregated.
\textbf{(b)} Composition of temporal annotations by annotation type (outer ring) and number of ground-truth moments per annotation unit (inner ring).
\textbf{(c)} Duration distribution of product evidence moments.}
    \label{fig:gpub_statistics}
\end{figure*}

As shown in Table~\ref{tab:benchmark_comparison}, GPUB distinguishes itself from existing benchmarks through its large product space and comprehensive evaluation of both product identification and temporal evidence localization. Figure~\ref{fig:gpub_statistics} further characterizes the internal data distribution of GPUB. Product evidence is both sparse and temporally distributed relative to the video context: videos are 4--5 minutes long on average, while 69.48\% of product evidence moments are shorter than 30 seconds. Moreover, 43.13\% of task-specific annotation units contain multiple ground-truth moments, with relevant evidence potentially occurring at multiple non-contiguous moments throughout a video. GPUB further provides text queries at different levels of specificity to capture product evidence localization at different granularities. Together, these characteristics require models not only to distinguish fine-grained products within a large product space, but also to localize and integrate temporally distributed evidence over extended video contexts.

\subsection{Evaluation Metrics}
\label{sec:evaluation_protocol}

\noindent
\textbf{Product Retrieval} is evaluated over the full catalog using
Recall@$K$, mean reciprocal rank (MRR), and normalized discounted
cumulative gain (nDCG@$K$), measuring top-$K$
retrieval success, the reciprocal rank of the target product, and
ranking quality with position discounting~\citep{voorhees2000trec,
jarvelin2002cumulated}. Only the target product is treated as relevant. \textbf{Product Moment Localization} is evaluated using temporal mAP
and Recall@$K$ at temporal intersection-over-union (tIoU) thresholds, with ranked predictions matched one-to-one to query-specific ground-truth moments. For \textbf{GPrU}, we introduce two joint metrics: Pair mAP for ranked product--moment predictions and Joint Recall for instance-level joint success, both defined below.
\par\smallskip

\noindent\textbf{Pair mAP.}
For pair-level evaluation, each predicted moment $\hat m_{ij}$ forms a product--moment pair $(p_i,\hat m_{ij})$. For benchmark instance $n$, pairs are processed in descending confidence order, and a pair is considered correct at threshold $\tau$ if $p_i=p_n^*$ and $\hat m_{ij}$ is greedily matched to the unmatched ground-truth moment $e\in\mathcal{E}_n^*$ with the highest tIoU, where $\mathrm{tIoU}(\hat m_{ij},e)\ge\tau$. We compute standard AP from the
ranked pairs for each instance and average it across all instances:
\begin{equation}
\operatorname{Pair\,mAP}@\tau
=
\frac{1}{N}\sum_{n=1}^{N}\operatorname{AP}_n^\tau.
\end{equation}
Unmatched predictions are treated as false positives, while unmatched ground-truth moments remain in the AP denominator. We report results at 
$\tau\in\{0.3,0.5\}$. 

\noindent\textbf{Joint Recall.}
Joint Recall measures instance-level joint success in product identification and evidence localization. An instance is counted as successful when the target product is ranked within the top $K$ and at least one predicted moment for that product overlaps a ground-truth evidence moment with $\mathrm{tIoU}\geq\tau$. The metric is computed as the fraction of successful instances over the benchmark. We report $K\in\{1,5\}$.

\section{Method}
\label{sec:method}
\subsection{Overview}
\begin{figure*}[t]
\centering
\includegraphics[width=0.85\textwidth]{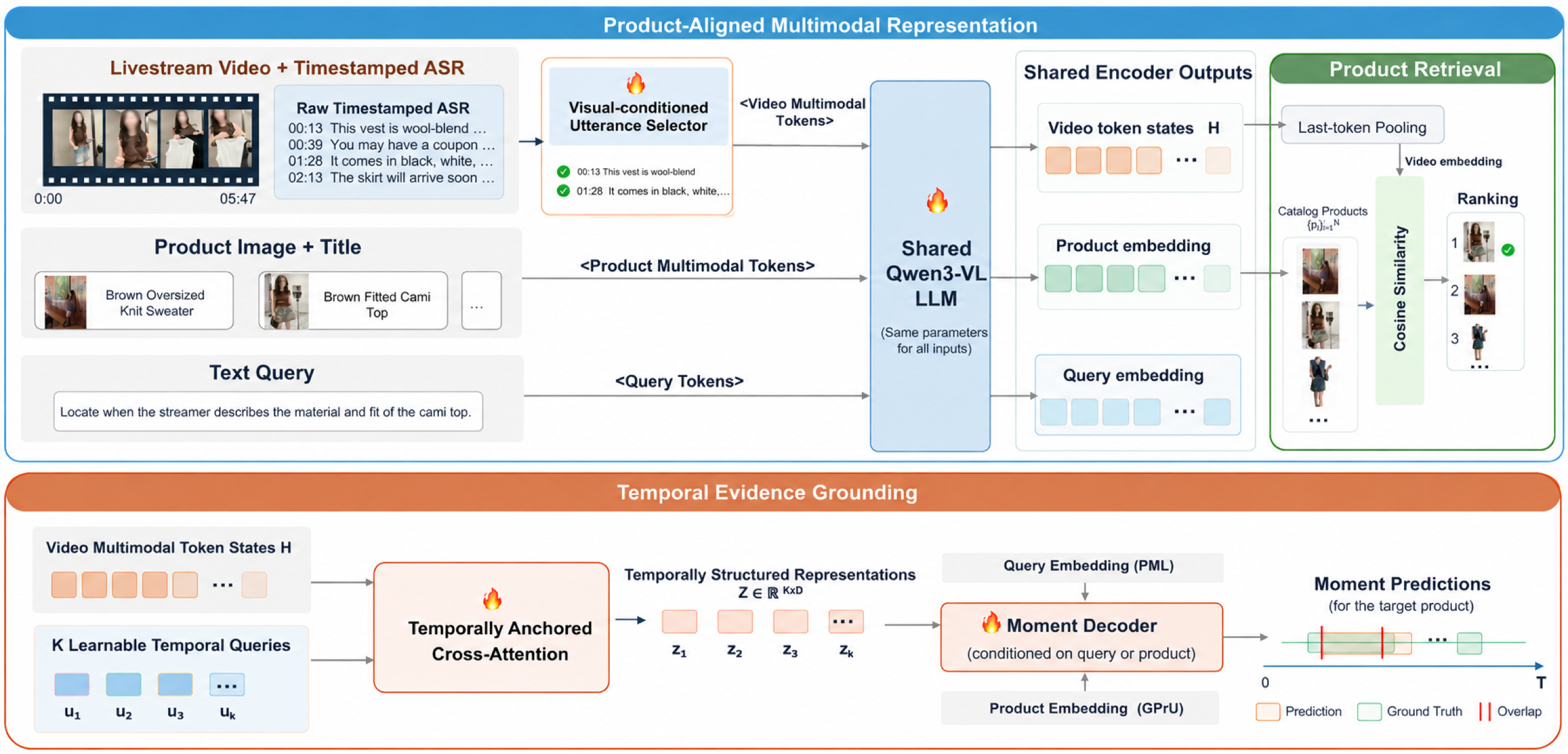}
\caption{\textbf{Overview of UniPro.} UniPro builds a shared product-aligned multimodal representation for product retrieval and temporal evidence grounding, where temporally structured representations are further constructed for moment localization.}
\label{fig:method_overview}
\end{figure*}

As illustrated in Fig.~\ref{fig:method_overview}, UniPro is built on Qwen3-VL-Embedding-2B and leverages shared multimodal representations to jointly perform product identification and temporal evidence localization. Given sampled video frames and selected timestamped ASR utterances, the shared encoder first produces multimodal token representations. These representations are used to derive a product-aligned video embedding for product retrieval and are further temporally structured for moment localization. Depending on the task, moment localization is conditioned on either a text query in Product Moment Localization or candidate products in GPrU.

\suppressfloats[t]

\subsection{Product-Aligned Multimodal Representation}

We first adapt Qwen3-VL-Embedding-2B to align livestream content with catalog products in a shared semantic space. Given a cropped product evidence moment and its corresponding catalog product, both are encoded by the shared backbone and optimized in a common embedding space using an in-batch contrastive objective. We then extend this product-level alignment from cropped evidence moments to complete livestream instances. Since livestream ASR is often lengthy and contains substantial content unrelated to the target product, encoding the full transcript consumes excessive tokens and introduces considerable redundancy. We therefore employ a lightweight Visual-Conditioned Utterance Selector that scores each timestamped utterance conditioned on a pooled visual representation and retains relevant utterances under a fixed token budget. The selected utterances preserve their original timestamps and temporal order. The selected ASR and sampled video tokens are jointly encoded by the product-aligned backbone, producing shared multimodal token states $H \in \mathbb{R}^{L\times d}$. For Product Retrieval, we normalize the last valid token state in $H$ as the global video embedding and rank catalog products by cosine similarity.

\subsection{Temporal Evidence Grounding}
\label{sec:method_training}
While the shared multimodal representation $H$ captures visual and spoken content, moment localization additionally requires organizing this information along the video timeline. We therefore apply Temporally Anchored Cross-Attention to obtain temporally structured representations from $H$. Specifically, $K$ learnable anchor queries distributed over the normalized video timeline attend to the multimodal tokens in $H$, producing $Z=\{z_k\}_{k=1}^{K}\in\mathbb{R}^{K\times d_t}$. During cross-attention, temporal information is incorporated through temporal encodings of both the multimodal tokens and anchor queries, together with a temporal-distance bias. Based on the temporal representations $Z$, we use a Moment Decoder to localize the evidence moments corresponding to a text query or candidate product. The Moment Decoder is implemented as a three-layer Transformer decoder with a set of learnable moment queries. The semantic condition is projected and added to the moment queries, which attend to $Z$ and jointly predict normalized temporal intervals and their confidence scores. For Product Moment Localization, the text-query embedding $q$ serves as the condition. For Grounded Product Understanding, each candidate-product embedding $p_i$ serves as the condition, with product--moment pairs ranked by combining the product matching score and moment confidence.

\section{Experiments}
\label{sec:experiments}

\subsection{Experimental Setup}
\label{sec:experimental_setup}

We evaluate two groups of representative baselines according to the requirements of the GPUB tasks. For Product Retrieval, we consider Vision--Language embedding models for efficient fine-grained matching between product evidence moments and catalog products. For Product Moment Localization and Grounded Product Understanding, we evaluate general-purpose MLLMs for temporal evidence localization over long-form livestreams. These baseline models and their detailed evaluation settings are provided in Appendix~\ref{app:evaluation_details}. UniPro is evaluated across all three tasks. Its implementation and training details, including training-data statistics and split integrity, are provided in Appendix~\ref{app:implementation}. All models are evaluated on the complete GPUB test set following the task-specific protocols in Section~\ref{sec:task_definition}.

\subsection{Main Benchmark Results}
\label{sec:main_results}

\begin{table*}[t]
    \centering
    \caption{Main results on the three GPUB tasks (\%). Metric definitions are provided in Section~\ref{sec:evaluation_protocol}. In panel~(c), Pair mAP and Joint R denote the product--moment correspondence metrics, while Ret. and Loc. denote product-retrieval and temporal-localization diagnostics, respectively. The subscript 30 indicates the fixed 30-candidate setting. Higher is better.}
    \label{tab:main_results}
    \scriptsize
    \setlength{\tabcolsep}{2.6pt}
    \renewcommand{\arraystretch}{0.94}

    \begin{tabular*}{\textwidth}{@{\extracolsep{\fill}}lcccccccc@{}}
        \toprule
        \multicolumn{9}{c}{\textit{\textbf{(a) Product Retrieval --- Full Catalog}}} \\
        \midrule
        \multirow{2}{*}{\textbf{Model}} &
        \multicolumn{4}{c}{\textbf{Recall}} &
        \multicolumn{3}{c}{\textbf{nDCG}} &
        \multirow{2}{*}{\textbf{MRR}} \\
        \cmidrule(lr){2-5}\cmidrule(lr){6-8}
        & \textbf{R@1} & \textbf{R@10} & \textbf{R@30} & \textbf{R@50}
        & \textbf{@10} & \textbf{@30} & \textbf{@50} & \\
        \midrule
        
        Chinese-CLIP-336px         & 1.73 & 6.35 & 10.27 & 13.22 & 3.72 & 4.65 & 5.20 & 3.23 \\
        GME-Qwen2-VL-2B            & 3.58 & 11.46 & 19.43 & 24.56 & 6.96 & 8.83 & 9.80 & 6.18 \\
        SigLIP2-SO400M-Patch14-384 & 3.85 & 11.56 & 18.43 & 22.08 & 7.30 & 8.93 & 9.61 & 6.48 \\
        FG-CLIP2-Base              & 4.22 & 11.51 & 16.96 & 19.71 & 7.48 & 8.76 & 9.28 & 6.62 \\
        Qwen3-VL-Embedding-2B      & 13.04 & 34.22 & 45.66 & 51.32 & 22.79 & 25.52 & 26.59 & 20.06 \\
        Qwen3-VL-Embedding-8B      & \underline{15.05} & \underline{37.89} & \underline{50.48} & \underline{57.02}
                                   & \underline{25.50} & \underline{28.48} & \underline{29.70} & \underline{22.54} \\
        \textbf{UniPro}        & \cellcolor{refblue}\textbf{33.47} & \cellcolor{refblue}\textbf{58.20} & \cellcolor{refblue}\textbf{69.11} & \cellcolor{refblue}\textbf{73.65}
                                   & \cellcolor{refblue}\textbf{45.31} & \cellcolor{refblue}\textbf{47.90} & \cellcolor{refblue}\textbf{48.75} & \cellcolor{refblue}\textbf{42.00} \\
        \midrule
    \end{tabular*}

    \vspace{0.10em}
    \begin{tabular*}{\textwidth}{@{\extracolsep{\fill}}lcccccccc@{}}

        \multicolumn{9}{c}{\textit{\textbf{(b) Product Moment Localization --- 2,497 Queries}}} \\
        \midrule
        \multirow{2}{*}{\textbf{Model}} &
        \multicolumn{4}{c}{\textbf{$\mathrm{tIoU}=0.3$}} &
        \multicolumn{4}{c}{\textbf{$\mathrm{tIoU}=0.5$}} \\
        \cmidrule(lr){2-5}\cmidrule(lr){6-9}
        & \textbf{R@1} & \textbf{R@5} & \textbf{R@10} & \textbf{mAP}
        & \textbf{R@1} & \textbf{R@5} & \textbf{R@10} & \textbf{mAP} \\
        \midrule
        Qwen2.5-VL-7B             & 14.71 & 20.10 & 22.61 & 17.26 & 6.17 & 7.85 & 8.76 & 6.96 \\
        Qwen3-VL-8B               & \underline{23.09} & \underline{33.49} & \underline{36.93} & \underline{27.96}
                                  & 10.89 & 15.43 & 17.16 & 12.96 \\
        InternVL3-8B              & 8.98 & 14.70 & 16.87 & 11.64 & 2.99 & 4.71 & 5.29 & 3.74 \\
        InternVideo2.5-7B         & 13.80 & 21.65 & 26.22 & 17.65 & 5.37 & 7.78 & 9.09 & 6.55 \\
        Keye-VL-1.5-8B            & 11.56 & 19.43 & 22.31 & 15.10 & 4.53 & 7.12 & 7.99 & 5.64 \\
        MiniCPM-V-4.5             & 6.27 & 11.26 & 12.65 & 8.43 & 1.91 & 3.40 & 3.71 & 2.56 \\
        GLM-4.6V-Flash            & 22.63 & 33.03 & 36.65 & 27.42
                                  & \underline{11.08} & \underline{16.12} & \underline{18.11} & \underline{13.31} \\
        Qwen3-VL-8B (SFT)         & 42.65 & 47.62 & 52.62 & 46.28 & 21.65 & 32.36 & 36.36 & 29.39 \\
        \textbf{UniPro}       & \cellcolor{refblue}\textbf{43.34} & \cellcolor{refblue}\textbf{64.05} & \cellcolor{refblue}\textbf{78.22} & \cellcolor{refblue}\textbf{53.50}
                                  & \cellcolor{refblue}\textbf{25.76} & \cellcolor{refblue}\textbf{38.20} & \cellcolor{refblue}\textbf{49.01} & \cellcolor{refblue}\textbf{32.80} \\
        \midrule
    \end{tabular*}

    \vspace{0.10em}
    \begin{tabular*}{\textwidth}{@{\extracolsep{\fill}}lcccccccc@{}}

        \multicolumn{9}{c}{\textit{\textbf{(c) Grounded Product Understanding --- 30 Candidates}}} \\
\midrule
\multirow{2}{*}{\textbf{Model}} &
\multicolumn{4}{c}{\textbf{Product--Moment Correspondence}} &
\multicolumn{4}{c}{\textbf{Diagnostics}} \\
\cmidrule(lr){2-5}\cmidrule(lr){6-9}
& \textbf{Pair mAP@.3} & \textbf{Pair mAP@.5} & \textbf{Joint R@1@.3} & \textbf{Joint R@5@.3}
& \shortstack{\textbf{Ret.}\\\textbf{R@1$_{30}$}} & \shortstack{\textbf{Ret.}\\\textbf{mAP$_{30}$}} & \shortstack{\textbf{Loc.}\\\textbf{R@1@.3}} & \shortstack{\textbf{Loc.}\\\textbf{mAP@.3}} \\
\midrule
Qwen2.5-VL-7B             & 3.53 & 1.57 & 4.13 & 7.07 & 19.20 & 27.82 & 10.47 & 11.22 \\
Qwen3-VL-8B               & \underline{10.13} & \underline{5.03} & \underline{18.03} & \underline{19.93} & \underline{39.10} & \underline{51.63} & \underline{15.92} & \underline{19.55} \\
InternVL3-8B              & 5.34 & 2.46 & 5.43 & 11.07 & 22.23 & 31.38 & 14.71 & 14.81 \\
InternVideo2.5-7B         & 0.44 & 0.07 & 1.23 & 2.20 & 11.63 & 16.86 & 1.58 & 5.26 \\
Keye-VL-1.5-8B            & 2.27 & 1.13 & 4.13 & 4.17 & 17.47 & 17.62 & 9.14 & 9.87 \\
MiniCPM-V-4.5             & 2.42 & 0.92 & 3.30 & 5.90 & 17.83 & 26.17 & 4.80 & 6.31 \\
GLM-4.6V-Flash            & 7.37 & 3.87 & 12.53 & 16.10
                            & 22.70 & 27.78 & 8.30 & 16.34 \\
Qwen3-VL-8B (SFT)         & 21.92 & 13.77 & 35.47 & 35.47 & 38.60 & 47.43 & 39.83 & 52.15 \\
\textbf{UniPro}       & \cellcolor{refblue}\textbf{24.58} & \cellcolor{refblue}\textbf{14.33} & \cellcolor{refblue}\textbf{38.81} & \cellcolor{refblue}\textbf{65.54}
                            & \cellcolor{refblue}\textbf{39.40} & \cellcolor{refblue}\textbf{52.54} & \cellcolor{refblue}\textbf{40.82} & \cellcolor{refblue}\textbf{56.76} \\
\bottomrule
\end{tabular*}
\end{table*}

Table~\ref{tab:main_results} summarizes the overall results. Results indicate that GPUB remains challenging for current models across all three tasks. On full-catalog Product Retrieval, even Qwen3-VL-Embedding-8B achieves only 15.05 R@1. The challenge extends to temporal grounding and becomes more pronounced in the joint GPrU setting, where Qwen3-VL-8B achieves only 10.13 Pair mAP@.3. Task-specific fine-tuning substantially improves its performance on both PML and GPrU, demonstrating the benefit of downstream supervision. UniPro further improves over this SFT baseline by 7.22 on PML mAP@.3 and 2.66 on GPrU Pair mAP@.3, reaching 53.50 and 24.58, respectively. On GPrU Pair mAP@.5, Qwen3-VL-8B (SFT) obtains 13.77, compared with 14.33 for UniPro. These results demonstrate the benefits of task-specific supervision and dedicated modeling across the complementary capabilities evaluated by GPUB. Additional results are provided in Appendix~A.3.

\subsection{Analysis}
\label{sec:benchmark_analysis}

\begin{figure*}[t]
\centering
\makebox[\textwidth][c]{\includegraphics[width=0.795\textwidth,trim=0 45 0 0,clip]{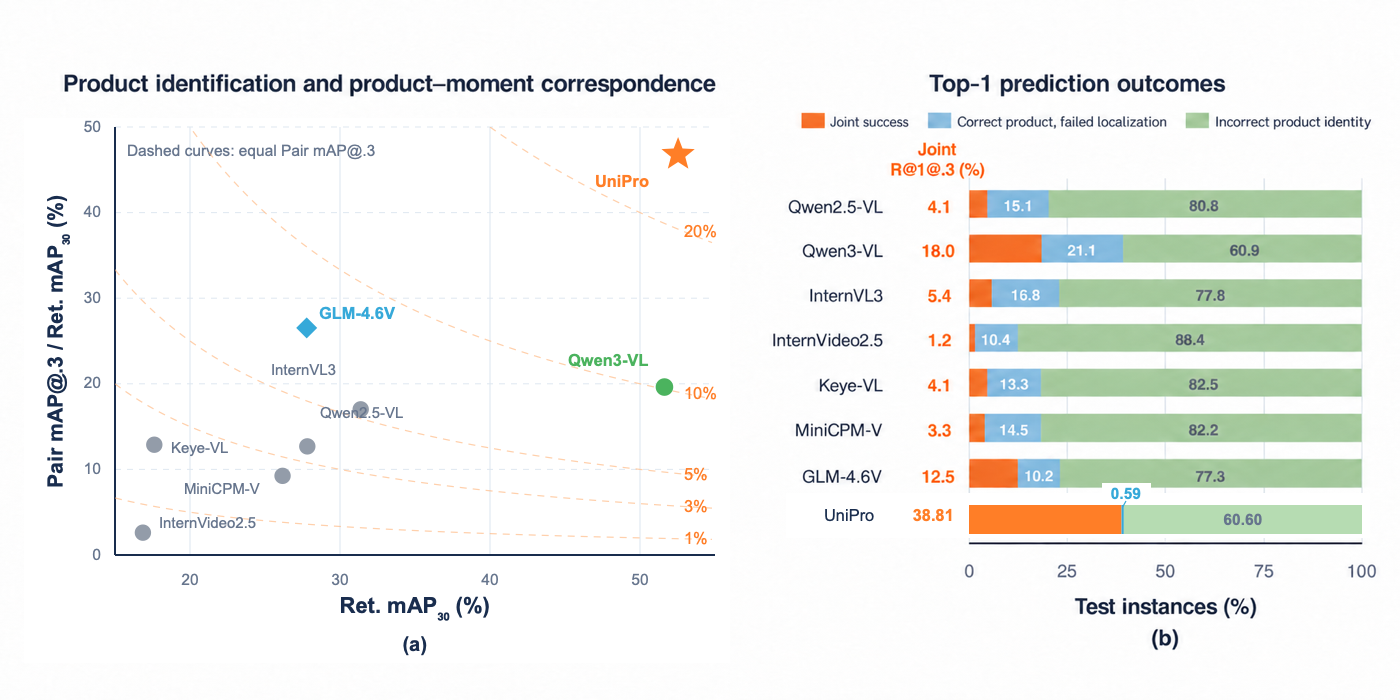}}\vspace{-4pt}
\caption{
\textbf{(a)} Comparison of product identification and product–moment correspondence across models. Ret. mAP$_{30}$ measures product identification, while normalized Pair mAP@.3 reflects product–moment correspondence; dashed curves indicate equal Pair mAP@.3. (b) Outcome decomposition of GPrU Top-1 predictions into joint success, localization failure after correct product identification, and incorrect product identity.}
\label{fig:product_moment_profiles}
\end{figure*}

\paragraph{Similar product identification performance but different GPrU performance.}
As shown in Fig.~\ref{fig:product_moment_profiles}(a),
strong product identification does not necessarily translate
into strong joint grounding performance. UniPro achieves a Ret. mAP$_{30}$
of 52.54 and a normalized Pair mAP@.3 of 46.78.
In the Top-1 comparison in Fig.~\ref{fig:product_moment_profiles}(b),
Qwen3-VL-8B and UniPro achieve nearly identical
Ret. R@1$_{30}$ scores of 39.10 and 39.40,
respectively, yet their Joint R@1@.3 scores differ substantially,
at 18.03 and 38.81.
For UniPro, the outcome decomposition comprises
38.81 joint success, 0.59 correct product identification with failed
localization, and 60.60 incorrect product identity.
This contrast highlights the need to jointly evaluate product identity
and its supporting temporal evidence.

\begin{table*}[!t]
\centering
\caption{Grounded Product Understanding ablations under the fixed 30-candidate protocol (\%).
\textbf{(a)} Representation architecture. \textbf{(b)} ASR content selection.}
\label{tab:task3_ablations}
\footnotesize
\setlength{\tabcolsep}{2.0pt}
\resizebox{\textwidth}{!}{%
\begin{tabular}[t]{lrrrrrr}
\multicolumn{7}{c}{\textbf{(a) Architecture design.}} \\[2pt]
\toprule
& \textbf{Ret.} & \multicolumn{2}{c}{\textbf{Loc. mAP}} & \textbf{Joint} & \multicolumn{2}{c}{\textbf{Pair mAP}} \\
\cmidrule(lr){3-4}\cmidrule(lr){6-7}
\textbf{Variant} & \textbf{R@1$_{30}$} & \textbf{@.3} & \textbf{@.5} & \textbf{R@1@.3} & \textbf{@.3} & \textbf{@.5} \\
\midrule
w/o Separate Retrieval Representation & 13.64 & 44.59 & 22.00 & 13.50 & 10.23 & 5.13 \\
\begin{tabular}[c]{@{}l@{}}w/o Temporally Structured\\Representations\end{tabular} & 38.80 & 50.66 & 28.75 & 37.94 & 21.57 & 11.60 \\
\textbf{Full UniPro} & \textbf{39.40} & \textbf{56.76} & \textbf{32.80} & \textbf{38.81} & \textbf{24.58} & \textbf{14.33} \\
\bottomrule
\end{tabular}%
\hspace{1.8em}%
\setlength{\tabcolsep}{2.8pt}%
\begin{tabular}[t]{lrrrrrr}
\multicolumn{7}{c}{\textbf{(b) ASR content selection.}} \\[2pt]
\toprule
& \textbf{Ret.} & \multicolumn{2}{c}{\textbf{Loc. mAP}} & \multicolumn{2}{c}{\textbf{Pair mAP}} & \textbf{Joint} \\
\cmidrule(lr){3-4}\cmidrule(lr){5-6}
\textbf{ASR policy} & \textbf{R@1$_{30}$} & \textbf{@.3} & \textbf{@.5} & \textbf{@.3} & \textbf{@.5} & \textbf{R@1@.3} \\
\midrule
No ASR & 37.97 & 48.70 & 26.83 & 19.94 & 10.33 & 36.47 \\
Random ASR & 39.23 & 54.36 & 32.04 & 22.57 & 12.85 & 38.27 \\
Unfiltered ASR & 38.20 & 56.12 & \textbf{33.52} & 23.76 & 13.92 & 37.64 \\
\textbf{Selected ASR} & \textbf{39.40} & \textbf{56.76} & 32.80 & \textbf{24.58} & \textbf{14.33} & \textbf{38.81} \\
\bottomrule
\end{tabular}%
}
\end{table*}

\paragraph{Fine-grained catalog ambiguity challenges product identification.}
We group samples by the similarity between each target product and its hardest negative, with results reported in Appendix Table~\ref{tab:gpru_difficulty}(a). Identification performance degrades as similarity increases: from low to high similarity, Ret. R@1$_{30}$ drops by 25.98, 9.76, and 27.15 for Qwen3-VL-8B, GLM-4.6V-Flash, and UniPro, respectively. The prevalence of visually and semantically similar products in GPUB makes fine-grained product identification particularly challenging. These errors also limit GPrU performance, since joint success requires both correct product identification and evidence localization.

\paragraph{Semantic and distributed evidence pose distinct challenges to temporal localization.}
As shown in Appendix Fig.~\ref{fig:query_type_dumbbell}, all seven general-purpose MLLMs perform better on fine-grained queries than on coarse-grained queries, suggesting that explicit attribute cues may facilitate temporal localization. This suggests that such cues provide anchors for temporal localization, whereas semantically defined evidence remains more challenging to ground. Meanwhile, GT-Recall@10 consistently decreases as the number of ground-truth moments increases (Appendix Table~\ref{tab:gpru_difficulty}(b)), indicating that evidence distributed across multiple moments is more difficult to recover completely.

\paragraph{Ablation Study.}
Table~\ref{tab:task3_ablations} examines representation design and ASR content selection. Using temporally structured representations for product retrieval instead of a separate retrieval representation reduces Ret. R@1$_{30}$ by 25.76 (Table~\ref{tab:task3_ablations}(a)), indicating that temporal structure is less suitable for product matching. Removing temporal structuring reduces Loc. mAP@.3 by 6.10 while leaving retrieval largely unchanged, confirming its importance for moment localization. For ASR selection (Table~\ref{tab:task3_ablations}(b)), removing ASR reduces Loc. mAP@.3 and Pair mAP@.3 by 8.06 and 4.64, respectively, showing that ASR provides useful evidence. Random and unfiltered ASR recover much of this loss. Compared with unfiltered ASR, selected ASR reduces the ASR token budget while maintaining comparable performance and achieving the highest Pair mAP@.3 (24.58). Overall, these results highlight the importance of task-specific representations and efficient ASR selection in UniPro.

\section{Conclusion}
\label{sec:conclusion}

We introduced GPUB, a large-scale benchmark for product understanding in livestream videos, centered on the new GPrU task of jointly identifying the target product and localizing its supporting temporal evidence. Our evaluation reveals persistent challenges arising from fine-grained catalog ambiguity, temporally distributed evidence, and the need to establish correct product--moment correspondence. We further develop UniPro, a unified baseline that addresses all three tasks within a shared framework. We hope GPUB will provide a foundation for future research on grounded multimodal understanding in realistic long-form videos.

\section*{AI Use Statement}
Generative AI tools were used for literature search, language editing, and polishing. All AI-assisted changes were reviewed and verified by the authors. The authors take full responsibility for the final content of this paper.

\section*{Project Page}
The anonymous project page and accompanying code are available at
\url{https://anonymous.4open.science/api/repo/gpub-evaluation-C2E4/file/project-page/index.html}.

\bibliography{iclr2027_conference}\bibliographystyle{iclr2027_conference}
\clearpage
\appendix
\section*{Appendix}

\section{Baseline Evaluation Details}
\label{app:evaluation_details}

This section describes the evaluation settings for the GPUB baselines,
including input construction, prompting, and model-output processing.

\subsection{Vision--Language Embedding Baselines}

For Product Retrieval, each annotated product evidence moment is represented
by 10 uniformly sampled RGB frames and the corresponding ASR transcript, while
each catalog product is represented by its main image and title. For each model, we use its officially released processor and follow the official
inference configuration for embedding extraction. On the video side, the
sampled frames and corresponding ASR are used to construct the query
representation; on the product side, the product image and title are used to
construct the catalog-product representation.

Qwen3-VL-Embedding-2B/8B~\citep{li2026qwen3vlembedding} and GME-Qwen2-VL-2B~\citep{zhang2024gme} directly encode
the corresponding multimodal inputs to produce the video and product
representations. For the dual-tower models, including FG-CLIP2-Base~\citep{xie2025fgclip2},
SigLIP2-SO400M-Patch14-384~\citep{tschannen2025siglip2}, and Chinese-CLIP-336px~\citep{yang2022chineseclip}, the visual and textual
inputs are encoded separately. On the video side, the visual representations
of the sampled frames are first mean-pooled and then fused with the
corresponding ASR representation. On the product side, the product image and
title representations are fused in the same manner. Both sides use a
visual--text fusion weight of $0.7{:}0.3$. The resulting video and product
representations are $\ell_2$-normalized, and products are ranked over the full
catalog using cosine similarity.

\subsection{Multimodal LLM Baselines}
\label{app:mllm_prompts}
\label{app:metric_details}

For a consistent comparison across MLLM baselines, including Qwen2.5-VL-7B~\citep{bai2025qwen25vl}, Qwen3-VL-8B~\citep{bai2025qwen3vl}, InternVL3-8B~\citep{zhu2025internvl3}, InternVideo2.5-7B~\citep{wang2025internvideo25}, Keye-VL-1.5-8B~\citep{keyevl2025}, MiniCPM-V-4.5~\citep{yu2025minicpmv45}, and GLM-4.6V-Flash~\citep{glm46vflash2025}, we use the same input preprocessing and inference settings. For each livestream video, we uniformly sample 64 frames and resize each frame so that its longer side is 448 pixels. The timestamped ASR is preserved in its original temporal order and truncated to at most 4,096 characters, with timestamps expressed relative to the beginning of the video. All models use greedy decoding and the same task-specific Chinese instructions. For Product Moment Localization, the instruction additionally provides the text query, whereas for Grounded Product Understanding, it provides the 30 candidate products through their titles and main images without an additional text query. For readability, the prompts are presented below in English translation.

\begin{center}
\setlength{\fboxsep}{7pt}
\fcolorbox{promptborder}{promptbg}{%
\parbox{0.92\linewidth}{%
\textbf{Prompt Box --- Product Moment Localization}\par\vspace{0.3em}
{\color{promptborder}\hrule height 0.4pt}\vspace{0.55em}
\small\raggedright

Given an e-commerce livestream video represented by uniformly sampled frames and
timestamped ASR, together with a text query:\par\medskip
\textbf{Video:}\par
\texttt{\{uniformly sampled video frames\}}\par\medskip
\textbf{Timestamped ASR:}\par
\texttt{\{timestamped ASR\}}\par\medskip
\textbf{Query:}\par\texttt{\{query\}}\par\medskip
Localize all temporal moments in the video that satisfy the query.\par\medskip
Assign an independent confidence score in $[0,1]$ to each predicted moment and rank the predictions by confidence.
Return at most 10 moments.\par\medskip

Output only structured JSON containing \texttt{moments}, where each moment is represented by its start and end timestamps and confidence score.

}}
\end{center}

\begin{center}
\setlength{\fboxsep}{7pt}
\fcolorbox{promptborder}{promptbg}{%
\parbox{0.92\linewidth}{%
\textbf{Prompt Box --- Grounded Product Understanding}\par\vspace{0.3em}
{\color{promptborder}\hrule height 0.4pt}\vspace{0.55em}
\small\raggedright

Given an e-commerce livestream video represented by uniformly sampled frames and
timestamped ASR, together with $N$ candidate products, each represented by its title and main image:\par\medskip
\textbf{Video:}\par
\texttt{\{uniformly sampled video frames\}}\par\medskip
\textbf{Timestamped ASR:}\par
\texttt{\{timestamped ASR\}}\par\medskip
\textbf{Candidate products:}\par
\texttt{\{}$(i,\text{title}_i,\text{image}_i)$\texttt{\}} for $i=1,\ldots,N$\par\medskip
Identify the candidate product being discussed in the video and localize all temporal moments that provide visual or spoken evidence about its attributes or characteristics.\par\medskip
Mere product appearances and segments primarily concerning promotion, price, inventory, or
casual conversation should not be considered evidence moments.\par\medskip
Return three ranked sets of predictions:\par
1. candidate products;\par 2. evidence moments;\par 3. product--moment pairs.\par\medskip
Assign an independent confidence score in $[0,1]$ to each prediction and rank each set by confidence.\par
Return at most 10 moments and 10 product--moment pairs.\par\medskip
Output only structured JSON containing \texttt{ranked\_products}, \texttt{moments}, and \texttt{pairs},
where each moment is represented by its start and end timestamps and each pair additionally contains
the corresponding candidate product index.

}}
\end{center}

\paragraph{Output Processing.}
For Product Moment Localization, the returned temporal intervals and their
confidence scores are used directly for temporal localization evaluation. For
Grounded Product Understanding, each model returns a ranked list of candidate
products, temporal intervals, and product--moment pair predictions. Product--moment predictions are ranked by the confidence scores generated by each
model for AP computation. Candidate indices are mapped back to their corresponding product IDs. Temporal
intervals are restricted to the current evaluation window and converted from
window-relative timestamps to absolute timestamps in the original video.
Invalid entries that cannot be parsed into the required format, contain
invalid candidate indices, or do not define valid temporal intervals are
discarded. If an output cannot be parsed, it is treated as an empty prediction.
For Pair mAP, at most the top 10 ranked product--moment pairs per instance are retained for all methods, matching the maximum number of ground-truth evidence moments in an instance.
The resulting predictions are evaluated
using the metrics defined in Section~\ref{sec:evaluation_protocol}.

\subsection{Supervised Fine-Tuning Results}
\label{app:sft_results}

To provide a controlled comparison under downstream supervision, we fine-tune representative MLLMs on the GPUB training split. Each model is trained following the GPrU formulation to jointly identify the target product and localize its supporting moments. The resulting models are evaluated on both PML and GPrU using the same evaluation protocols as in the main experiments. Table~\ref{tab:sft_results} reports the detailed results. Beyond the main metrics discussed in Sec.~\ref{sec:main_results}, UniPro shows larger gains at higher recall cutoffs on PML, indicating better coverage of relevant evidence moments. On GPrU, improvements in both product identification and evidence localization diagnostics suggest that its gain in joint product--moment prediction is supported by both capabilities.

\begin{table*}[t]
\centering
\caption{\textbf{Supervised fine-tuning results on GPUB (\%).}
\textbf{(a)} Product Moment Localization.
\textbf{(b)} Grounded Product Understanding under the fixed 30-candidate protocol.
The best SFT baseline in each column is underlined; UniPro is highlighted in blue.}
\label{tab:sft_results}
\scriptsize
\setlength{\tabcolsep}{2.6pt}
\renewcommand{\arraystretch}{0.94}

\begin{tabular*}{\textwidth}{@{\extracolsep{\fill}}lrrrrrrrr@{}}
\toprule
\multicolumn{9}{c}{\textit{\textbf{(a) Product Moment Localization}}} \\
\midrule
\multirow{2}{*}{\textbf{Model}} &
\multicolumn{4}{c}{\textbf{$\mathrm{tIoU}=0.3$}} &
\multicolumn{4}{c}{\textbf{$\mathrm{tIoU}=0.5$}} \\
\cmidrule(lr){2-5}\cmidrule(lr){6-9}
& \textbf{R@1} & \textbf{R@5} & \textbf{R@10} & \textbf{mAP}
& \textbf{R@1} & \textbf{R@5} & \textbf{R@10} & \textbf{mAP} \\
\midrule
Qwen2.5-VL-7B (SFT) & 34.23 & 41.84 & 46.38 & 36.17 & 16.53 & 25.22 & 29.16 & 21.16 \\
Keye-VL-1.5-8B (SFT) & 32.81 & 41.37 & 46.50 & 35.05 & 16.09 & 25.42 & 29.64 & 21.08 \\
GLM-4.6V-Flash (SFT) & 40.28 & 44.48 & 47.13 & 42.05 & 21.61 & \underline{32.93} & 35.63 & 28.10 \\
Qwen3-VL-8B (SFT) & \underline{42.65} & \underline{47.62} & \underline{52.62} & \underline{46.28} & \underline{21.65} & 32.36 & \underline{36.36} & \underline{29.39} \\
\midrule
\textbf{UniPro} & \cellcolor{refblue}\textbf{43.34} & \cellcolor{refblue}\textbf{64.05} & \cellcolor{refblue}\textbf{78.22} & \cellcolor{refblue}\textbf{53.50}
& \cellcolor{refblue}\textbf{25.76} & \cellcolor{refblue}\textbf{38.20} & \cellcolor{refblue}\textbf{49.01} & \cellcolor{refblue}\textbf{32.80} \\
\bottomrule
\end{tabular*}

\vspace{0.8em}

\begin{tabular*}{\textwidth}{@{\extracolsep{\fill}}lrrrrrrrr@{}}
\toprule
\multicolumn{9}{c}{\textit{\textbf{(b) Grounded Product Understanding --- 30 Candidates}}} \\
\midrule
\multirow{2}{*}{\textbf{Model}} &
\multicolumn{4}{c}{\textbf{Product--Moment Correspondence}} &
\multicolumn{4}{c}{\textbf{Diagnostics}} \\
\cmidrule(lr){2-5}\cmidrule(lr){6-9}
& \textbf{Pair mAP@.3} & \textbf{Pair mAP@.5} & \textbf{Joint R@1@.3} & \textbf{Joint R@5@.3}
& \shortstack{\textbf{Ret.}\\\textbf{R@1$_{30}$}} & \shortstack{\textbf{Ret.}\\\textbf{mAP$_{30}$}} & \shortstack{\textbf{Loc.}\\\textbf{R@1@.3}} & \shortstack{\textbf{Loc.}\\\textbf{mAP@.3}} \\
\midrule
Qwen2.5-VL-7B (SFT) & 12.68 & 8.54 & 20.34 & 23.16 & 20.71 & 27.51 & 31.42 & 41.37 \\
Keye-VL-1.5-8B (SFT) & 11.47 & 7.63 & 18.21 & 21.26 & 18.92 & 21.34 & 30.16 & 39.82 \\
GLM-4.6V-Flash (SFT) & 18.74 & 13.26 & 24.86 & 31.82 & 29.63 & 31.37 & 36.72 & 47.36     \\
Qwen3-VL-8B (SFT) & \underline{21.92} & \underline{13.77} & \underline{35.47} & \underline{35.47} & \underline{38.60} & \underline{47.43} & \underline{39.83} & \underline{52.15} \\
\midrule
\textbf{UniPro} & \cellcolor{refblue}\textbf{24.58} & \cellcolor{refblue}\textbf{14.33} & \cellcolor{refblue}\textbf{38.81} & \cellcolor{refblue}\textbf{65.54}
& \cellcolor{refblue}\textbf{39.40} & \cellcolor{refblue}\textbf{52.54} & \cellcolor{refblue}\textbf{40.82} & \cellcolor{refblue}\textbf{56.76} \\
\bottomrule
\end{tabular*}
\end{table*}
\FloatBarrier

\subsection{Difficulty Analysis}
\label{app:difficulty_analysis}

\paragraph{GPrU difficulty factors.}
As shown in Table~\ref{tab:gpru_difficulty}, we examine two sources of difficulty in GPrU: product ambiguity and temporal evidence complexity. The product-ambiguity analysis uses the subset of examples with valid hardest-negative similarity scores. Product identification degrades as the target product becomes more similar to its hardest negative, while temporal localization becomes more difficult as the number of evidence moments increases.

\begin{table}[H]
\centering
\caption{\textbf{Analysis of product identification and temporal localization under different difficulty factors (\%).}
\textbf{(a)} Product identification performance across different levels of similarity between the target product and its hardest negative.
\textbf{(b)} Temporal localization performance across different numbers of ground-truth moments.
GT-Recall@10 measures the recall of annotated moments among the top 10 predictions for the ground-truth product at tIoU $=0.3$.}
\label{tab:gpru_difficulty}
\scriptsize
\setlength{\tabcolsep}{2.6pt}
\renewcommand{\arraystretch}{0.92}

\begin{tabular*}{0.90\textwidth}{@{\extracolsep{\fill}}lrrrrrrr@{}}
\toprule
\multicolumn{8}{c}{\textbf{\textit{(a) Product Ambiguity}}} \\
\midrule
\multicolumn{2}{l}{\textbf{Similarity}} &
\multicolumn{3}{c}{\textbf{Ret. R@1$_{30}$}} &
\multicolumn{3}{c}{\textbf{Joint R@1@.3}} \\
\cmidrule(lr){3-5}\cmidrule(lr){6-8}
& & \textbf{Qwen3-VL} & \textbf{GLM-4.6V} & \textbf{UniPro}
  & \textbf{Qwen3-VL} & \textbf{GLM-4.6V} & \textbf{UniPro} \\
\midrule
\multicolumn{2}{l}{Low} & 52.54 & 35.35 & 67.97 & 7.03 & 12.11 & 65.04 \\
\multicolumn{2}{l}{Medium} & 39.65 & 28.91 & 52.73 & 6.64 &  9.96 & 50.59 \\
\multicolumn{2}{l}{High} & 26.56 & 25.59 & 40.82 & 3.71 & 11.52 & 38.48 \\
\midrule
\multicolumn{8}{c}{\textbf{\textit{(b) Temporal Evidence Complexity}}} \\
\midrule
\multirow{2}{*}{\textbf{GT moments}} &
\multirow{2}{*}{\textbf{$N$}} &
\multicolumn{3}{c}{\textbf{GT-Recall@10}} &
\multicolumn{3}{c}{\textbf{Pair mAP@.3}} \\
\cmidrule(lr){3-5}\cmidrule(lr){6-8}
& & \textbf{Qwen3-VL} & \textbf{GLM-4.6V} & \textbf{UniPro}
  & \textbf{Qwen3-VL} & \textbf{GLM-4.6V} & \textbf{UniPro} \\
\midrule
1        & 1,127 & 49.59 & 33.19 & 91.75 & 17.12 & 12.03 & 28.37 \\
2        &   731 & 34.43 & 23.59 & 83.79 &  8.99 &  7.02 & 23.83 \\
$\geq 3$ & 1,142 & 18.11 & 11.95 & 67.53 &  3.96 &  2.99 & 21.32 \\
\midrule
Overall  & 3,000 & 33.92 & 22.77 & 80.59 & \textbf{10.13} & \textbf{7.37} & \textbf{24.58} \\
\bottomrule
\end{tabular*}
\end{table}
\FloatBarrier

\paragraph{Query-type breakdown.}
The Product Moment Localization evaluation set contains 1,300 coarse-grained
queries with 1,703 ground-truth moments and 1,197 fine-grained queries
with 1,506 ground-truth moments. Figure~\ref{fig:query_type_dumbbell} compares
the two query types for the seven general-purpose MLLMs using mAP@0.3.

\begingroup
\setlength{\intextsep}{3pt}
\setlength{\abovecaptionskip}{3pt}
\setlength{\belowcaptionskip}{0pt}
\begin{figure}[H]
  \centering
  \includegraphics[width=0.78\linewidth]{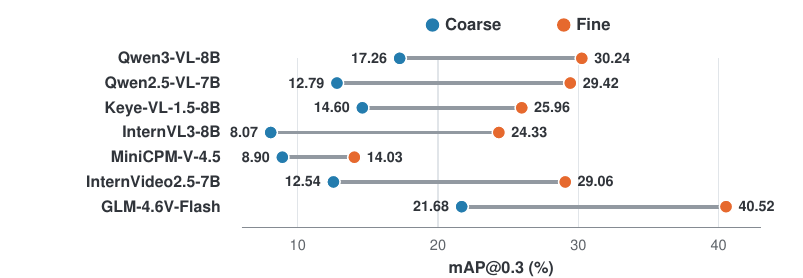}
  \vspace{-4pt}
  \caption{Product Moment Localization performance by query type for seven general-purpose MLLMs, measured by mAP@0.3.}
  \label{fig:query_type_dumbbell}
\end{figure}
\endgroup
\FloatBarrier

\section{UniPro Implementation Details}
\label{app:implementation}

\subsection{Architecture Details}

\paragraph{Utterance Selector.}
Each timestamped utterance is encoded into a representation \(u_i\), augmented with its start time, end time, and duration, while the video is summarized into a global visual representation \(v\). The selector computes the selection score as follows:
\begin{equation}
s_i^{\mathrm{sel}}=\mathrm{MLP}\!\left([u_i;\,v;\,u_i\odot v;\,|u_i-v|]\right).
\end{equation}
Here, $\odot$ denotes element-wise multiplication and $|u_i-v|$ the element-wise absolute difference; the four components are concatenated before scoring. Higher-scoring utterances are retained as complete units, preserving their timestamps and temporal order.

\paragraph{Temporal Evidence Grounding.}
Video tokens are associated with frame timestamps, while ASR tokens inherit their utterance intervals; both are represented by normalized temporal center and duration. The number of uniformly spaced temporal anchors is determined by video duration $T$ as $K(T)=\operatorname{clip}(\lceil T/(5\,\mathrm{s})\rceil,16,128)$. Temporal encodings are added to both the anchor queries and timestamped multimodal tokens, while a temporal-distance bias between each anchor center and token interval is applied to the cross-attention logits. Cross-attention then yields $K$ temporally structured representations, one per anchor. The moment decoder uses $N_m=10$ queries per semantic condition, distinct from the $K$ temporal anchors. For GPrU, all candidate products share the same temporal representation $Z$, avoiding repeated video encoding. Product--moment pairs $(p,j)$ are ranked by $s_{p,j}=\log\operatorname{softmax}_{p}(s_p^{\mathrm{ret}}/0.07)+\log\sigma(c_{p,j})$, where $s_p^{\mathrm{ret}}$ is the product retrieval similarity, $c_{p,j}$ is the moment confidence logit, and the softmax is over candidate products.

\subsection{Training Strategy}
\label{app:training_strategy}

UniPro is trained in three stages using the datasets described below.

\paragraph{Stage I: Video--Product Alignment.}
We adapt Qwen3-VL-Embedding-2B with LoRA using an in-batch InfoNCE objective:
\begin{equation}
\mathcal{L}_{\mathrm{S1}}
=
-\frac{1}{B}\sum_{i=1}^{B}
\log
\frac{\exp(\operatorname{sim}(v_i,p_i)/\tau)}
{\sum_{j=1}^{B}\exp(\operatorname{sim}(v_i,p_j)/\tau)}.
\label{eq:stage1_objective}
\end{equation}
Here, $v_i$ and $p_i$ are the embeddings of a video and its matched product, $B$ is the contrastive batch size, and $\operatorname{sim}$ denotes cosine similarity. The matched product is the positive, and the remaining products in the batch serve as negatives.

\paragraph{Stage II: ASR-Aware Multimodal Adaptation.}
Stage II initializes from the product-aligned model learned in Stage I, while the Qwen3-VL-Embedding-2B backbone remains frozen. The utterance selector is optimized on timestamped livestream inputs using the video--product matching objective, without utterance-level supervision. Hard utterance selection is applied in the forward pass, with a straight-through gate providing surrogate gradients to the selector and a soft budget loss regularizing the retention ratio.

\paragraph{Stage III: Temporal Evidence Grounding.}
During Stage III, the multimodal encoder remains frozen while the temporal grounding modules are optimized; the selector is updated after an initial warm-up. Predicted moments are assigned to ground-truth moments by Hungarian matching with
\begin{equation}
\mathcal{C}_{\mathrm{match}}
=
0.5\mathcal{C}_{\mathrm{conf}}
+
5\mathcal{C}_{1}
+
2\mathcal{C}_{\mathrm{tGIoU}},
\label{eq:stage3_matching}
\end{equation}
where the terms measure confidence, normalized temporal-boundary distance, and temporal generalized IoU, respectively. The training objective is
\begin{equation}
\begin{aligned}
\mathcal{L}_{\mathrm{S3}} ={}&
3\mathcal{L}_{\mathrm{conf}}
+
5\mathcal{L}_{1}
+
2\mathcal{L}_{\mathrm{tGIoU}}
+
0.3\mathcal{L}_{\mathrm{rank}} \\
&+
\lambda_d(t)
\left[
\mathcal{L}_{\mathrm{fg}}
+
0.1\left(
\mathcal{L}_{\mathrm{start}}
+
\mathcal{L}_{\mathrm{end}}
\right)
\right].
\end{aligned}
\label{eq:stage3_objective}
\end{equation}

$\mathcal{L}_{1}$ measures boundary error on the normalized timeline, and $\mathcal{L}_{\mathrm{tGIoU}}=1-\operatorname{mean}(\mathrm{tGIoU})$. Confidence targets gradually transition from matching labels to localization-quality scores, defined as the maximum IoU between each proposal and any ground-truth moment. The pairwise ranking loss $\mathcal{L}_{\mathrm{rank}}$ encourages higher-quality proposals to receive higher confidence, weighting each pair by its quality gap. For the auxiliary dense supervision, \(\mathcal L_{\mathrm{fg}}\) uses weighted binary cross-entropy with soft targets given by the overlap ratio between each temporal slot and the ground-truth intervals, while \(\mathcal L_{\mathrm{start}}\) and \(\mathcal L_{\mathrm{end}}\) use Gaussian heatmaps centered on the nearest ground-truth boundaries. 

\subsection{GPUB Training Split and Split Integrity}
The GPUB training split is constructed using the same verification, evidence-annotation, and query-construction pipeline as the evaluation set. It contains 6,145 livestream windows spanning 465.29 hours, and the training candidate catalog contains 136,180 products. Detailed statistics are reported in Table 6. To prevent data leakage, windows sharing the same source livestream or target product are grouped and assigned to the same split, ensuring that neither source livestreams nor target products overlap across the training, validation, and test sets.

\begingroup
\setlength{\intextsep}{6pt}
\begin{table}[H]
    \centering
    \caption{GPUB split statistics. PML includes coarse- and fine-grained localization queries.}
    \label{tab:unipro_split_statistics}
    \small
    \setlength{\tabcolsep}{3.5pt}
    \begin{tabular}{lrrr}
        \toprule
        \textbf{Statistic} & \textbf{Train} & \textbf{Validation} & \textbf{Test} \\
        \midrule
        Livestream windows       & 6,145  & 500   & 3,000 \\
        Source livestreams       & 2,311  & 239   & 1,798 \\
        Unique target products   & 2,693  & 214   & 1,641 \\
        Product evidence moments & 15,708 & 1,373 & 8,975 \\
        PML queries              & 5,393  & 433 & 2,497 \\
        \bottomrule
    \end{tabular}
\end{table}
\endgroup

\subsection{Implementation and Optimization Settings}

The implementation and optimization settings are specified separately for the three training stages.

\textbf{Stage I} is trained for 8,000 steps on 4 GPUs with a per-GPU batch size of 8 and gradient accumulation of 4, using AdamW with a learning rate of $10^{-4}$, LoRA rank 8, scaling factor $\alpha=32$, and InfoNCE temperature $\tau=0.1$. \textbf{Stage II} is trained for 2,000 steps using the same learning rate as Stage I. \textbf{Stage III} is trained for 6,000 steps on 2 GPUs with a per-GPU batch size of 8 and a learning rate of $5\times10^{-5}$. We uniformly sample 64 frames from each video. ASR selection is performed in 30-s blocks with 3--8 utterances per block and an overall 2,048 token ASR budget. We use AdamW with cosine learning-rate decay to 0.1 times the initial rate, 200 warm-up steps, weight decay of 0.01, and a maximum gradient norm of 1.0. The selector begins updating after 500 steps. The ranking loss forms ordered pairs with a quality gap above 0.1 and uses a confidence-logit margin of 0.2, while the boundary heatmaps use $\sigma=1$ temporal slot. 

\section{Ethics and Data Governance}
\label{app}
GPUB is constructed from real-world e-commerce livestream data. To protect privacy, the original audio tracks are removed from all benchmark videos, and spoken content is provided only as de-identified timestamped ASR. Faces of livestream hosts and other identifiable individuals are blurred. Visible host nicknames and account IDs, QR codes, contact details, and store names are masked or replaced. Viewer account identifiers appearing in video frames are detected using OCR and pixelated. Personal names and nicknames in ASR are replaced, while personally identifiable or sensitive information in product images, titles, and attributes is removed or obscured. Internal storage paths and original livestream and product identifiers are replaced with benchmark-specific identifiers, and timestamps are expressed relative to each video window.  The released data have undergone privacy review and will be distributed under specified access and licensing conditions.

\end{document}